\documentclass[10pt, conference, compsocconf]{IEEEtran}
\usepackage{fancyhdr} % Required for custom headers
\usepackage{lastpage} % Required to determine the last page for the footer
\usepackage{extramarks} % Required for headers and footers
\usepackage[usenames,dvipsnames]{color} % Required for custom colors
\usepackage{graphicx} % Required to insert images
\usepackage{listings} % Required for insertion of code
\usepackage{courier} % Required for the courier font
\usepackage{lipsum} % Used for inserting dummy 'Lorem ipsum' text into the template
\usepackage{caption}
\usepackage{pdfpages}
\usepackage{graphicx}
\usepackage{caption}
\usepackage{subcaption}
\usepackage{setspace}
\usepackage{amsmath}
\usepackage{algorithm}
\usepackage[noend]{algpseudocode}
\usepackage{amsmath}
\usepackage{epstopdf}
\newcommand\norm[1]{\left\lVert#1\right\rVert}

\begin{document}
%
% paper title
% can use linebreaks \\ within to get better formatting as desired
\title{Deep Recurrent Neural Networks for Sequential Phenotype Prediction in Genomics}

\author{\IEEEauthorblockN{Farhad Pouladi}
\IEEEauthorblockA{Faculty of Applied Sciences\\ of Communications,Tehran, Iran\\
UMTS Project Partners bv. (SPAA05),\\ Waalwijk, Netherlands\\
pouladi@ictfaculty.ir}
\and
\IEEEauthorblockN{Hojjat Salehinejad}
\IEEEauthorblockA{Farnoud Data Communication Company,\\ Kerman, Iran\\ hojjat.salehinejad@uoit.net}
\and
\IEEEauthorblockN{Amir Mohammad Gilani}
\IEEEauthorblockA{Mobile Communication Company\\ of Iran (MCI), Tehran, Iran\\
am.gilani@mci.ir}}
% make the title area
\maketitle

\begin{abstract}
In analyzing of modern biological data, we are often dealing with ill-posed problems and missing data, mostly due to high
dimensionality and multicollinearity of the dataset. In this paper, we have proposed a system based on matrix factorization (MF) and deep recurrent neural networks (DRNNs) for genotype imputation and phenotype sequences prediction. In order to model the long-term dependencies of phenotype data, the new Recurrent Linear Units (ReLU) learning strategy is utilized for the first time. The proposed model is implemented for parallel processing on central processing units (CPUs) and graphic processing units (GPUs). Performance of the proposed model is compared with other training algorithms for learning long-term dependencies as well as the sparse partial least square (SPLS) method on a set of genotype and phenotype data with 604 samples, 1980 single-nucleotide polymorphisms (SNPs), and two traits. The results demonstrate performance of the ReLU training algorithm in learning long-term dependencies in RNNs. 

\end{abstract}


\IEEEpeerreviewmaketitle

\section{Introduction}
% no \IEEEPARstart

Machine learning is a practical approach to deal with real world challenges such as to model neck pain and motor training induced plasticity \cite{baarbe1}, \cite{baarbe2}. Such methods have been widely used to solve difficult optimization problems \cite{h1}, \cite{h2}. Opposition based learning is one example which has achieved successful results in medical image processing and optimization problems \cite{sh1}, \cite{sh2}, \cite{h4}. The genome-wide association (GWA) studies have discovered many convincingly replicated associations for complex human diseases using high-throughput single-nucleotide polymorphism (SNP) genotypes \cite{nature2}, \cite{EuroGen}.
The genotype imputation has been used for fine-map associations and facilitates the combination of results across studies \cite{nature2}. 
The issue of missing genotype data and its imputation implies creating individualistic genotype data \cite{EuroGen}. Impact of even small amounts of missing data on a multi-SNP analysis is of great importance
for the complex diseases research \cite{EuroGen}. There are several programs such as BEAGLE \cite{t1}, MaCH \cite{t2}, and IMPUTE2 \cite{t3},
which provide imputation capability of untyped variants. 

The sparse partial least squares (SPLS) and least absolute shrinkage and selection operator (LASSO) methods are well-known for simultaneous dimension reduction and
variable selection \cite{spls}, \cite{lasso}. 
The LASSO is a shrinkage and selection method for linear regression, which attempts to minimize an error function. This function is typically the sum of squared errors with a bound on the sum of the absolute values of the coefficients \cite{lasso}. 
The partial least squares (PLS) regression is used as an alternative approach to the ordinary least squares (OLS) regression method \cite{spls}. The SPLS method is the sparse version of PLS method, which simultaneously works to 
achieve good predictive performance and variable selection by producing sparse linear combinations of the original predictors \cite{spls}. 

In general, matrix factorization is a technique to decompose a matrix for multivariate data into two matrices with $F$ latent features \cite{InfT}. Many matrix factorization techniques have been proposed to increase its performance, such as non-negative \cite{nature1},
sparse \cite{cibc1}, non-linear \cite{cibc3}, and kernel-based approaches \cite{cibc}.
A kernel non-negative matrix factorization method is proposed for feature extraction and classification
of micro-array data in \cite{cibc}. Performance evaluation of this method for eight different gene samples has showed better performance over linear as well as other well-known kernel-based matrix factorization approaches.

\begin{figure*}[!tp]
\centering
\includegraphics[scale=0.7]{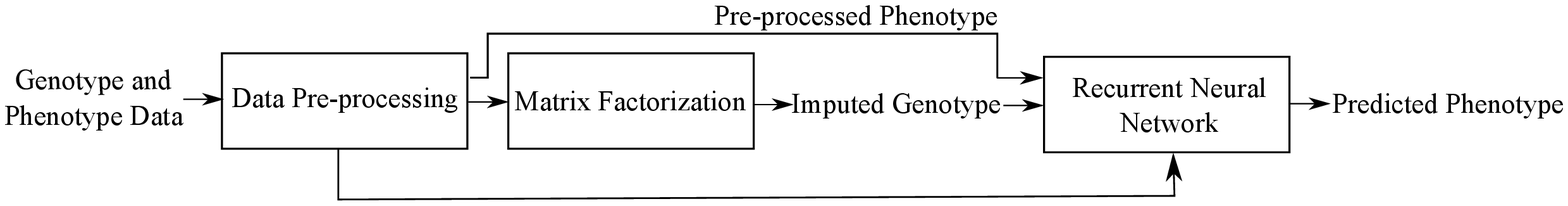}
  \caption{The proposed system model.}
  \label{fig:system_model}
\end{figure*}
\begin{figure*}[!tp]
\centering
\includegraphics[scale=0.8]{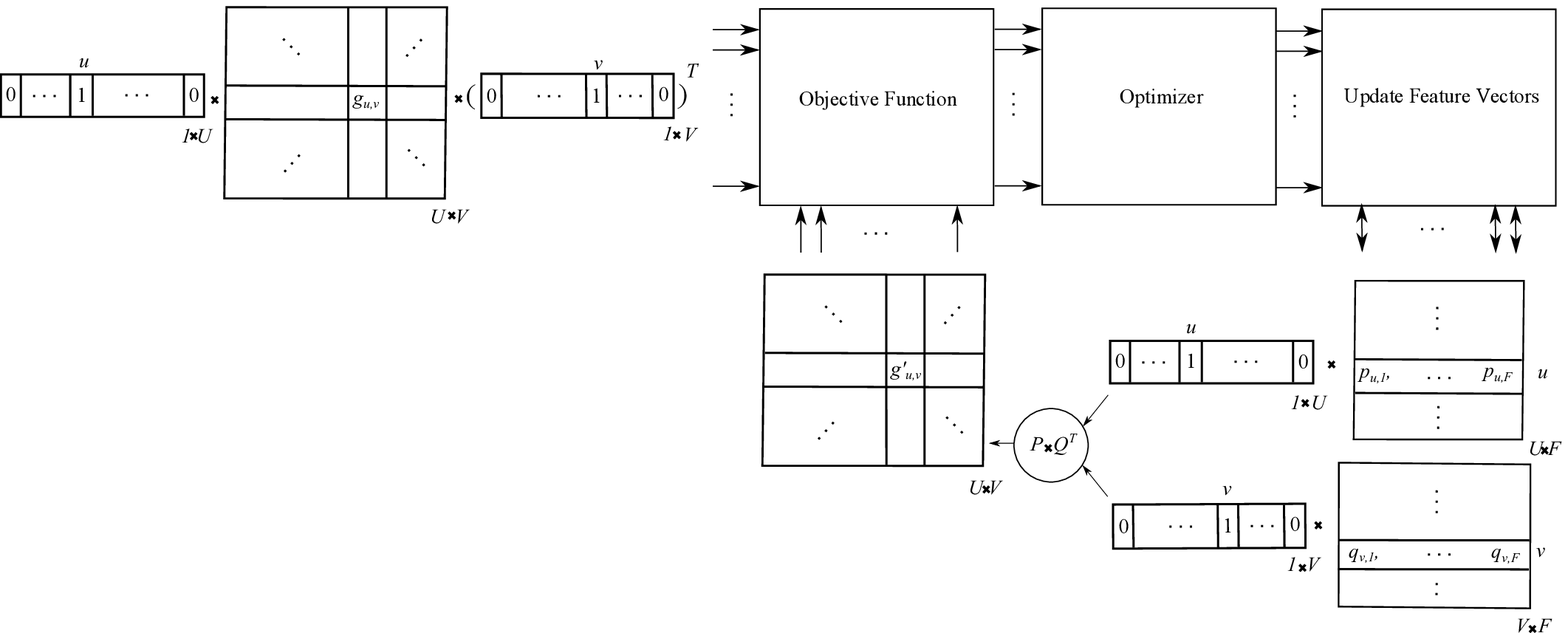}
  \caption{The matrix factorization model.}
  \label{fig:mf_model}
\end{figure*}

A sparse matrix factorization method has been proposed for tumor classification using gene expression data, \cite{cibc1}. In this approach, the gens are selected using a sparse matrix factorization method and then the features are extracted to be fed into 
a support vector machine (SVM) for tumor samples classification. It is reported that the performance results have been improved versus the non-sparse matrix factorization techniques.
The artificial neural network (ANN) is another successful machine learning approach for prediction and classification applications \cite{h5}. As an example, a feed-forward ANNs
model and a Bayesian approach are utilized to impute missing genotype data of SNPs in \cite{EuroGen}. Sequence modelling is one of the most areas in machine learning. This is due to the fact that a large class of phenomenal and data around us is made of sequences of data with particular patterns. Some examples are retail data, speed recognition, natural language modelling, music generation and genotype data for medical applications. With the great practical advances in deep learning, this state-of-the-art machine learning technique is the key for many problems in science and engineering. Recurrent neural network (RNN), due to its recurrent connections is considered as a subcategory of deep learning methods. This powerful model is capable of learning temporal patters and in sequential data. The power of RNN arises from its hidden state, which works as the ``memory'' of system to remember past important features for the future decision makings. The hidden state is consisted of high-dimensional non-linear dynamics which enables modelling any phenomena, if trained well \cite{Boulanger_2012}, \cite{Le_2015}. 

In this paper, we are proposing a new model for missing genotype imputation and phenotype prediction using matrix factorization and RNNs. In this model, a simple but efficient matrix factorization method is used for missing genotype imputation. Then, the imputed genotypes are used along the sequence of available phenotype data to train our RNN with the recently developed ReLU learning approach. In order to evaluate performance of the ReLU approach in learning long-term dependencies in phenotype data, it is compared with the LSTM-RNN and SRNN approaches. 

In the next section, the data structure of the dataset used for the experiments is described. In Section III, the methodologies based on the matrix factorization technique and DRNN is discussed in detail. The experimental results as well are comparative analysis are provided in Section IV. 
Finally, the paper is concluded in Section V and some guidelines are further developments are provided.

\section{Data Structure}

For the experiments, we are using a set of data provided for our research by Afzalipour research hospital.
The genotype data contains genotypes of 1980 SNPs for 604
observations and the phenotype data provides measurements of two phenotype, called trait 1 and trait 2. 
Out of 1980 SNPs provided in the
genotype data, 5$\%$ contain missing genotypes. The percentage of observations with missing genotypes for each SNP varies
from 1 to 25. For each trait, 30 randomly selected observations have missing values.

\section{Methodology}

In order to deal with the missing genotype and phenotype problem, we are utilizing the matrix factorization (MF) and RNNs techniques to fit prediction models as in Figure \ref{fig:system_model}. 
To do so, after data pre-processing, the genotype dataset with missing values is imported into the MF system to predict the missing genotype values. By having the estimated genotype dataset and 
corresponding phenotypes, the RNN is utilized in a supervised manner to train a network model for prediction of phenotypes, based on the known genotype-phenotype pairs. 
Each stage is described in details in the following subsections.

\begin{figure*}[!tp]
\footnotesize
\centering
        \begin{subfigure}[b]{0.28\textwidth}
           \centering 
                 \includegraphics[width=1\linewidth]{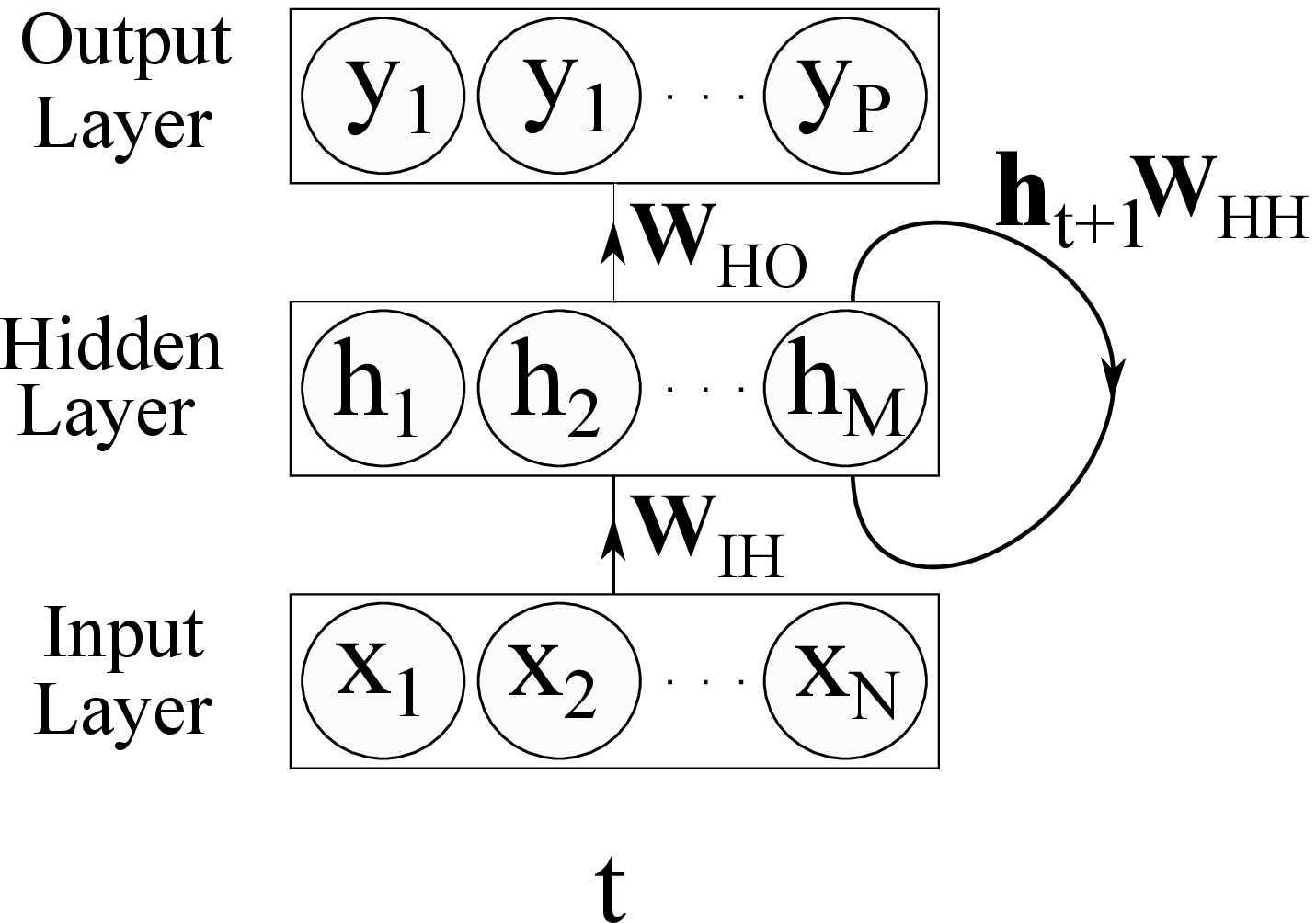}
                \caption{Fodled SRNN.}
                \label{fig:srnn_folded}
        \end{subfigure}%    
~  
 \begin{subfigure}[b]{0.45\textwidth}
    \centering 
                 \includegraphics[width=1\linewidth]{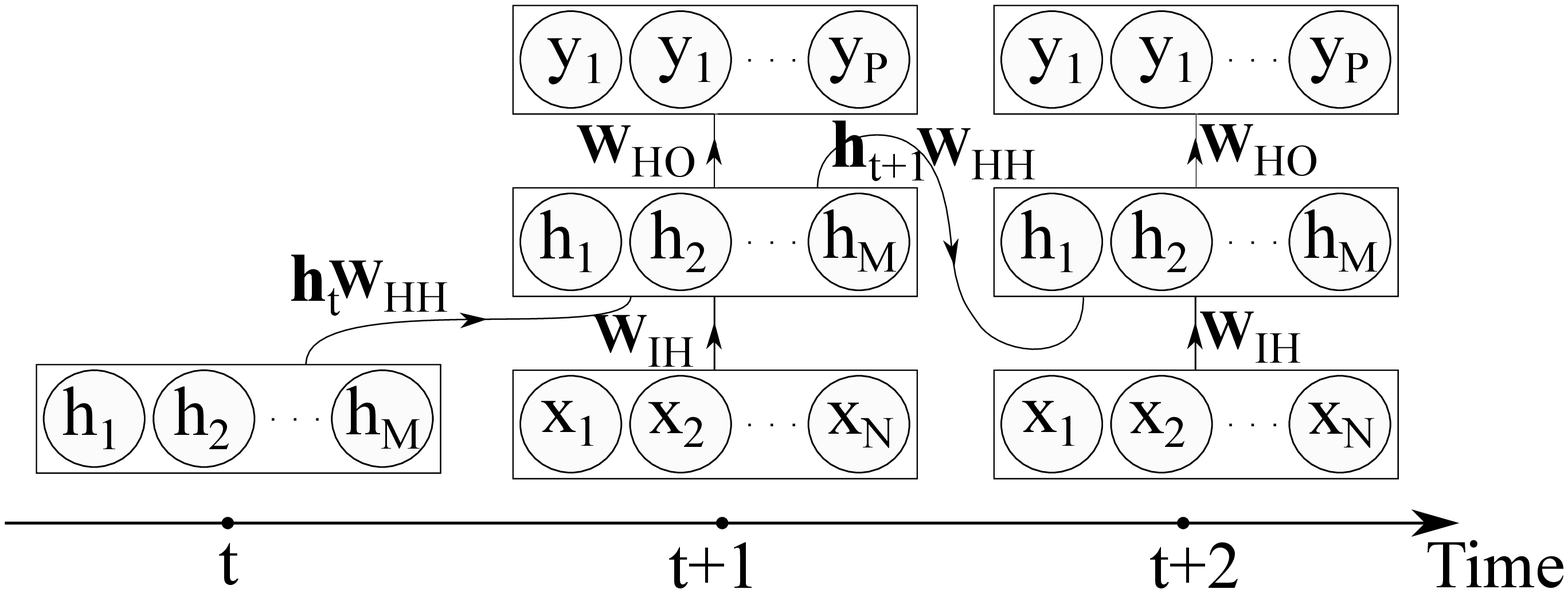}
                \caption{Unfolded SRNN through time.}
                \label{fig:srnn_unfolded}
        \end{subfigure}%
~
        \caption{A simple recurrent neural network (SRNN) and its unfolded structure through time. To keep the figure simple, biases are not shown.}
                \label{fig:srnn}     
\end{figure*}

\subsection{Data Pre-processing}
In general, the SNP genotypes (AA, BB, AB, or Null) are denoted with integer numbers for computational purposes, however, some programs may be able to work with this AA/AB/BB format directly. The data pre-processing step
is an opportunity to clean data, remove noise, and translate the genotype data and indicate/distinguish missing values from the available data. 

\subsection{Matrix Factorization Model}

The proposed MF structure for genotype data imputation is presented in Figure \ref{fig:mf_model}. In this model we consider $U$ number of samples and $V$ number of SNPs. Therefore, the genotype data 
is structured as a $U \times V$ matrix, called $G_{U \times V}$.   
The objective of MF technique is to estimate two matrices, $P_{U \times F}$ and $Q_{V \times F}$  with $F$ latent features, such that their product $G_{U \times V}^{'}$ estimates $G_{U \times V}$ as:

\begin{equation}
G \approx G^{'}=P \times Q^{T},
\end{equation}
where each element of the genotype matrix $G^{'}$ is computed by using the dot product such as:

\begin{equation}
 g^{'}_{u,v}=\sum_{f=1}^{F}p_{u,f}q_{f,v}.
\end{equation}

In order to find the best values for the matrices $P$ and $Q$, we need to minimize the objective function which describes the difference between the $G$ and $G^{'}$ genotype matrices \cite{InfT}. To do so, the gradient descent algorithm is
utilized as the optimizer in Figure \ref{fig:mf_model} to update the feature matrices $P$ and $Q$ iteratively. 

The above procedure is illustrated in pseudocode as in Algorithm \ref{psudocode}. As it is demonstrated, the parameters are set in the initialization step and random values in range $[a,b]$ 
are allocated to the feature matrices $P$ and $Q$, \cite{InfT}. Based on the availability of each genotype such as $G_{u,v}\neq \emptyset \enspace$ for all $\enspace \{u,v\}\in\{\{1,...,U\},\{1,...,V\}\}$, the estimated genotype matrix $G_{u,v}^{'}$
is computed. The objective function is then formulated with respect to $P$ and $Q$ as:

\begin{equation}
 min \enspace e(G,P,Q)^{2}=\sum_{u=1}^{U} \sum_{v=1}^{V}(G_{u,v}-G_{u,v}^{'})^2+\frac{\beta}{2}(\norm{P}_{F}+\norm{Q}_{F})
\end{equation}
where Forbenius norms of $P$ and $Q$ are used for regularization under control of parameter $\beta$ to prevent over-fitting of model by penalizing it with extreme parameter values \cite{InfT}.  
Normally $\beta$ is set to some values in the range of 0.02, such that $P$ and $Q$ can approximate $G$ without having to contain large numbers. The feature matrices of $P$ and $Q$ are updated as:

\begin{equation}
 P^{updated}=P-\alpha \frac{\partial e^{2}}{\partial P}
\end{equation}
and 
\begin{equation}
 Q^{updated}=Q-\alpha \frac{\partial e^{2}}{\partial Q}
\end{equation}
respectively, where $\alpha$ represents the learning rate and is practically set to $0.0001$.

%\makeatletter
%\def\BState{\State\hskip-\ALG@thistlm}
%\makeatother

\begin{algorithm}[t]
\caption{Matrix factorization}\label{psudocode}
\begin{algorithmic}[1]
\State $\textbf{Initialization}$
\State $P \gets rand([a,b])$
\State $Q \gets rand([a,b])$
\For {each epoch}
\If {$G_{u,v}\neq \emptyset \enspace \forall \enspace \{u,v\}\in\{\{1,...,U\},\{1,...,V\}\}$}
\State{$G_{u,v}^{'} \gets P\times Q^{T}$}
\State{$e^{2} \gets \sum_{u=1}^{U} \sum_{v=1}^{V}(G_{u,v}-G_{u,v}^{'})^2+\frac{\beta}{2}(\norm{P}_{F}+\norm{Q}_{F})$}
\State $P \gets P-\alpha \frac{\partial e^{2}}{\partial P}$
\State $Q \gets Q-\alpha \frac{\partial e^{2}}{\partial Q}$
\EndIf
\EndFor

\end{algorithmic}
\end{algorithm}

\subsection{Recurrent Neural Network with Rectified Linear Unit Model}

The utilized RNN in the proposed model in Figure~\ref{fig:system_model} is consisted of input, hidden, and output layers, where each layer is consisted of corresponding units. The input layer is consisted of $N$ input units, where its inputs are defined as a sequence of vectors through time $t$ such as 
$\{..., \textbf{x}_{t-1}, \textbf{x}_{t}, \textbf{x}_{t+1}...\}$ where $\textbf{x}_{t}=(x_{1}, x_{2}, ..., x_{N})$. In a fully connected  RNN, the inputs units 
are connected to hidden units in the hidden layer, where the connections are defined with a weight matrix $\textbf{W}_{IH}$. The hidden layer is consisted of $M$ hidden units $\textbf{h}_{t}=(h_{1}, h_{2}, ..., h_{M})$, which are connected to each other through time with recurrent connections. As it is demonstrated in Figure~\ref{fig:srnn_unfolded}, the hidden units are initiated before feeding the inputs. The hidden layer structure defines the state space or ``memory'' of the system, defined as 

\begin{equation}
\textbf{h}_{t} = f_{H}(\textbf{W}_{IH}\textbf{x}_{t}+\textbf{W}_{HH}\textbf{h}_{t-1}+\textbf{b}_{h})
\label{eq:SRNN_hidden_state}
\end{equation}
where $f_{H}(.)$ is the hidden layer activation function and $\textbf{b}_{h}$ is the bias vector of the hidden units. The hidden units are connected to the output layer with weighted connections $\textbf{W}_{HO}$. The output layer has $P$ units such as $\textbf{y}_{t}=(y_{1}, y_{2}, ..., y_{P})$ which are estimated as 
\begin{equation}
\textbf{y}_{t} = f_{O}(\textbf{W}_{HO}\textbf{h}_{t}+\textbf{b}_{0})
\label{eq:SRNN_outcome}
\end{equation}
where $f_{O}(.)$ is the output layer activations functions and $\textbf{b}_{0}$ is the bias vector.  

Learning long term dependencies in RNNs is a difficult task \cite{Le_2015}. This is due to two major problems which are vanishing gradients and exploding gradients. The long-short-term memory (LSTM) method is one of the popular methods to overcome the vanishing gradient problem. A recent proposed method suggests that proper initialization of the RNN weights with rectified linear units has good performance in modeling long-range dependencies \cite{Le_2015}. In this approach, the model is trained by utilizing back-propagation through time (BPTT) technique to compute the derivatives of error with respect to the weights. The reported performance analysis show that this method has comparable results in comparision to the LSTM method, with much less complexity. 

\begin{figure}[t]
\centering
\includegraphics[scale=0.3]{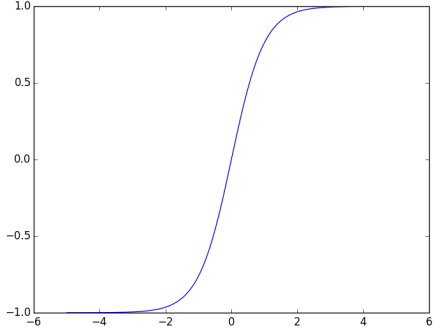}
  \caption{The ``tanh'' activation function.}
  \label{fig:tanh}
\end{figure}

\begin{figure}[t]
\centering
\includegraphics[scale=0.3]{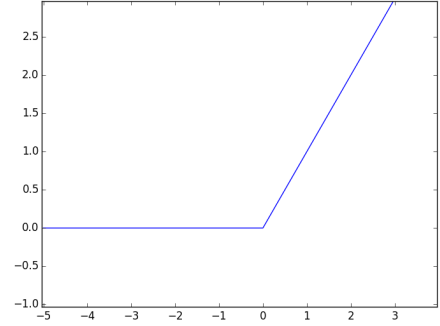}
  \caption{The ``rectified linear unit (ReLU)'' activation function. }
  \label{fig:relu}
\end{figure}

In this model, each new hidden state vector is inherited from the previous hidden vector by copying its values, adding the effects of inputs, and finally, replacing negative state values by zero. In other words, this means that the recurrent weight matrix is initialized to an identity matrix and the biases are set to zero. This procedure is in fact replacing the "tanh" activation function (Figure \ref{fig:tanh}) with a rectified linear unit (ReLU)(Figure \ref{fig:relu}). The ReLU in fact is modelling the behaviour of LSTM. In LSTM, the gates are set in a way that there is no decay to model long-term dependencies. In ReLU, when the error derivatives for the hidden units are back-propagated through time they remain constant provided no extra error-derivatives are added \cite{Le_2015}. 


\section{Experimental Results}
The proposed model is implemented for parallel processing using Theano in Python \cite{python}, \cite{Scipy}, \cite{Theano}. In this section, we are presenting the performance result from comparison between simple RNN, LSTM, and ReLU training methods for phenotype sequence prediction. The ReLU method is then compared with the well-known sparse partial least square (SPLS) method.

\subsection{Parameter Setting}

\begin{table}[t]
\centering
\footnotesize
\caption{Parameter setting for the experiments.}
\begin{tabular}{|c|c|c|}
\hline
Parameter Name & Parameter Descritption & Parameter Value \\ \hline
$\alpha$         & Learning Rate          & 0.001           \\ \hline
$\beta$          & Regularization Control & 0.02            \\ \hline
$N_{EP}$       & Number of Epochs       & 5000            \\ \hline
$U$              & Number of Samples      & 604             \\ \hline
$V$              & Number of SNPs         & 1980            \\ \hline
$F$           & Number of Features      & 400               \\ \hline
$eta_{1}$        & $eta$ for Trait 1     & 0.2               \\ \hline
$eta_{2}$           & $eta$ for Trait 2       & 0.2              \\ \hline
$K_{1}$        & $K$ for Trait 1     & 3               \\ \hline
$K_{2}$           & $K$ for Trait 2       & 4             \\ \hline
$[a,b]$           & Boundary values for $P$ and $Q$    & [0,1]            \\ \hline
\end{tabular}
\label{T:parameter}
\end{table}

As it is described in Section II, it is assumed that the genotypes data is either missing
or observed. Therefore, the observed genotypes are represented as $G_{u,v}\in \{0,1,2\}$ and the missing (null) data is represented as $G_{u,v}=5$ for the experiments \cite{nature2}. 
The genotype dataset $G_{U,V}$ is consisted of $V=1980$ SNPs for $U=604$ observations. 
Parameter setting for all the experiments are presented in Table \ref{T:parameter} adapted from the literature, [14], [19], [29], unless a
change is mentioned. As recommended in the literature, the 10-fold cross-validation is used to tune 
the parameters $eta$ and $K$ for the SPLS algorithm using the SPLS 
package in R programming language \cite{R}, \cite{splsR}, \cite{P_spls}, \cite{spls}. The parameter tuning is conducted separately for each 
provided trait in the phenotype dataset for $1\leq K \leq 10$. The optimal values are provided in Table \ref{T:parameter}. 

Due to small size of data samples, $\%80$ of available genotype and phenotype data is used for training the ANN, $\%10$ for validation, and $\%10$ for testing. Regarding the SPLS algorithm, $\%80$ of the provided data is considered as training data and $\%20$ as test data.

\subsection{Simulation Results Analysis}

In this subsection, performance results of the proposed method is presented and compared with the SPLS method for the described dataset in Section II. 
Performance of the methods is evaluated by measuring the correlation between the original genotype and phenotype values with the corresponding predicted values. 

\begin{figure}[t!]
    \centering
    \footnotesize
    \begin{subfigure}[t]{0.5\textwidth}
        \centering
               \includegraphics[scale=0.37]{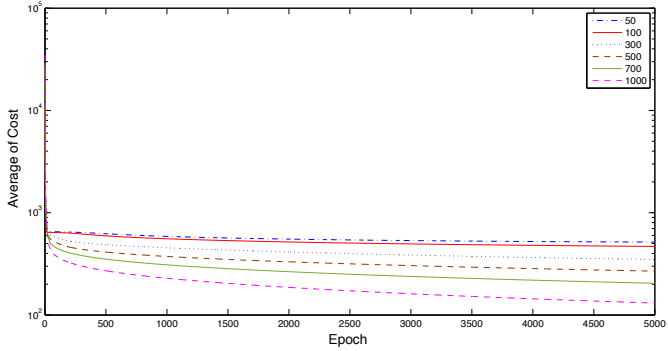}
        \caption{Logarithmic view for average of costs.}
    \end{subfigure}%
     
    \begin{subfigure}[t]{0.5\textwidth}
        \centering
        \includegraphics[scale=0.51]{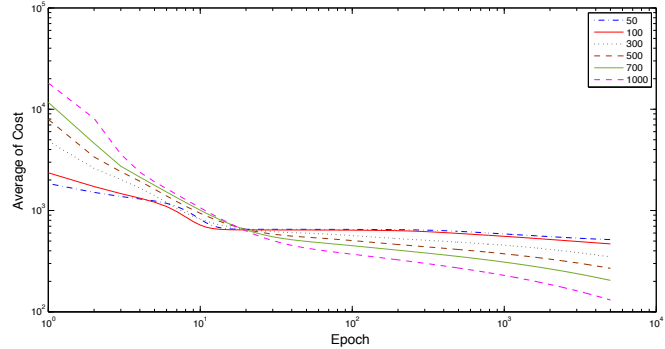}
        \caption{Logarithmic view for average of costs and epochs.}
    \end{subfigure}
    \caption{Average of costs versus epochs for genotype imputation using matrix factorization technique with different number of latent features.}
    \label{F:costMF} 
\end{figure}

\begin{figure}[t]
\centering
\includegraphics[scale=0.5]{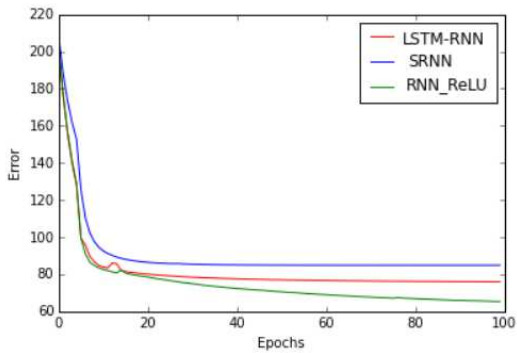}
  \caption{Training error comparison between SRNN, LSTM-RNN, and ReLU-RNN.}
  \label{fig:sim_rnn}
\end{figure}
The average of cost for fitting the MF model for different epochs is presented in Figure \ref{F:costMF}. In order to have a deeper look in details, the figure is presented in linear and logarithmic scales.
The measure to compute the average of cost is the mean-square error, which is the average of squared difference between targets and the output of the MF model. As the results show, the more number of latent 
features $F$ results in less average of cost. In addition, the model is trying to fit better that before in each epoch, however, after epoch 110 the progress is not very significant. The performance results for the 
success percentile in missing genotype data and success percentile in the whole genotype data construction is provided in Table \ref{Tab:MF_Res}.
There is a trade-off between the number of features and performance. This is due to the fact that increasing the number of features increases the computational complexity.

In Table~\ref{Tab:ANN_Res}, performance  of the ReLU-RNN for the genotype prediction is presented. As it is showed, the ReLU method has better training performance comparing to the results of the SPLS method in Table~\ref{T:spls}. The missing genotype represents the error for the test dataset in percentile while the original genotype represents the training error values in percentile. Since the training dataset has been seen by the model during training, it is reasonable to see that performance of the methods for the training dataset is better than the unseen test data.

In Figure~\ref{fig:sim_rnn}, the training error of the SRNN, LSTM-RNN, and ReLU-RNN algorithms are compared. For better illustration, the first 100 training epochs are presented. As the results show, the SRNN algorithm has more training error than the LSTM-RNN method. This is while the ReLU-RNN approach has the least training error comparing to LSTM-RNN. At the early epochs, we see that the LSTM and ReLU are almost at the same training loss, however, the ReLU achieves less error in further epochs. 
\begin{table}[t]
\centering
\footnotesize
\caption{Performance results of the matrix factorization technique success for different latent features $F$.}
\begin{tabular}{|c|c|c|}
\hline
F    & Missing Genotype  & Genotype Matrix Construction \\ \hline
50   &  $\%$52.45                       &  $\%$71.47                                   \\ \hline
100  &  $\%$65.16                       &  $\%$76.11                                   \\ \hline
300  &  $\%$67.53                       &  $\%$88.11                                   \\ \hline
500  &  $\%$68.80                       &  $\%$95.72                                    \\ \hline
700  &  $\%$70.88                       &  $\%$96.91                                    \\ \hline
1000 &  $\%$72.48                       &  $\%$97.56                                   \\ \hline
\end{tabular}
\label{Tab:MF_Res}
\end{table}
\begin{table}[t]
\centering
\footnotesize
\caption{Performance results of the ReLU-RNN for successful phenotype traits prediction.}
\begin{tabular}{|c|c|c|c|l}
\cline{1-4}
Trait & Mode    & Original Genotype& Missing Genotype&  \\ \cline{1-4}
      &  Train	&  $\%$82.75                     &  $\%$80.53                        &  \\ \cline{2-4}
1     &  Validation	&  $\%$86.56                    &  $\%$80.22                        &  \\ \cline{2-4}
      &  Test	&  $\%$84.66                     &  $\%$80.25                        &  \\ \cline{1-4}
      &     Train     &  $\%$80.62                     &  $\%$75.01                        &  \\ \cline{2-4}
2     &  Validation	&  $\%$85.25               &  $\%$80.54                       &  \\ \cline{2-4}
      &  Test	&  $\%$79.58                    &  $\%$78.29                      &  \\ \cline{1-4}
\end{tabular}
\label{Tab:ANN_Res}
\end{table}
\begin{table}[!t]
\centering
\footnotesize
\caption{Performance results in terms of correlation between the predicted and original phenotype data for the sparse partial least squares (SPLS) method.}
\begin{tabular}{|c|c|c|c|l}
\cline{1-4}
Trait & Mode    & Original Phenotype & Missing Phenotype &  \\ \cline{1-4}
1      &  Train	&  $\%$75.10                     & $\%$69.52                        &  \\ \cline{2-4}    
      &  Test	& $\%$54.56                      & $\%$51.53                         &  \\ \cline{1-4}
2      &Train     & $\%$80.53                     & $\%$72.85                        &  \\ \cline{2-4}  
      &  Test	&  $\%$69.82                    & $\%$56.42                       &  \\ \cline{1-4}
\end{tabular}
\label{T:spls}
\end{table}

\section{Conclusion and Future Works}
In this paper, a novel model is proposed which utilizes matrix factorization and deep recurrent neural networks (DRNN) for genotype imputation and phenotype sequences prediction. Sine we are interested in keeping track of sequences with long-term dependencies in genomics, the state-of-the-art recited linear unit learning method is used. 

The performance results show the with the ReLU methods has a better performance in training comparing to the LSTM-RNN and simple RNN methods. The ReLU learning methods also has less computational complexity comparing to the LSTM method. For future research, it is interesting to analyze other recent advances in deep learning for genotype-phenotype application; particularly that these algorithms are moving toward more simple designs which is suitable for big data application.

% that's all folks
\end{document}